\documentclass[runningheads]{llncs}

\usepackage{graphicx}
\usepackage{multirow}
\usepackage[hidelinks]{hyperref}

\begin{document}
\title{KERMIT -- A Transformer-Based Approach for Knowledge Graph Matching}

\author{Sven Hertling\inst{1}\thanks{The authors contributed equally to this paper.}\orcidID{0000-0003-0333-5888} \and
Jan Portisch\inst{1,2}$^\star$\orcidID{0000-0001-5420-0663} \and
Heiko Paulheim\inst{1}\orcidID{0000-0003-4386-8195}}

\authorrunning{Sven Hertling, Jan Portisch, Heiko Paulheim}

\institute{Data and Web Science Group, University of Mannheim, Germany\\
\email{\{sven,jan,heiko\}@informatik.uni-mannheim.de}\\
\and
SAP SE, Walldorf, Germany\\
\email{jan.portisch@sap.com}}

\maketitle
\begin{abstract}
One of the strongest signals for automated matching of knowledge graphs and ontologies are textual concept descriptions. 
With the rise of transformer-based language models, text comparison based on meaning (rather than lexical features) is available to researchers. However, performing pairwise comparisons of all textual descriptions of concepts in two knowledge graphs is expensive and scales quadratically (or even worse if concepts have more than one description). 
To overcome this problem, we follow a two-step approach: we first generate matching candidates using a pre-trained sentence transformer (so called \emph{bi-encoder}). In a second step, we use fine-tuned transformer cross-encoders to generate the best candidates. We evaluate our approach on multiple datasets and show that it is feasible and produces competitive results.

\end{abstract}

\section{Introduction}
\emph{Ontology Matching} (OM) is the task of finding correspondences between classes, properties, and instances of two or more ontologies. 
Multiple techniques exist to perform the matching operation algorithmically~\cite{euzenat_ontology_2013_ch_4}. 
Labels and descriptions are among the strongest signals concerning the semantics of an element of a knowledge graph (KG). 
Here, matcher developers often borrow strategies from the natural language processing (NLP) community to determine similarity between two strings.
Since the attention mechanism~\cite{DBLP:conf/nips/VaswaniSPUJGKP17} has been presented, so called transformer models gained a lot of traction in the NLP area, and transformers achieved remarkable results on tasks such as machine translation~\cite{DBLP:conf/nips/VaswaniSPUJGKP17} or question answering~\cite{DBLP:conf/naacl/DevlinCLT19,DBLP:journals/corr/abs-1910-03771}.
Traditional transformer models use a cross-encoder which requires that two sentences are used as input to predict the target variable. Since this does not scale when a lot of comparisons have to be performed, blocking methods are typically used to reduce the search space for true positives. However, this approach potentially sacrifices recall since traditional blocking methods rely on basic comparisons such as string overlap. With such blocking methods, a lot of useful correspondences are not found. 
Sentence BERT~\cite{DBLP:conf/emnlp/ReimersG19} (SBERT) overcomes this disadvantage by providing an approach that allows to derive embeddings for sentences such that two texts are close in this space when they share the same meaning. The idea is to train two transformer models simultaneously with a siamese network architecture. In this paper, we evaluate an SBERT-based approach for ontology and KG matching. A challenge of matching ontologies with transformers is the fact that cross-encoders typically have to be fine-tuned; however, this process requires the existence of a (partial) reference alignment, which is not always accessible. Thus, we also use a rather simple matcher which provides a high precision (with a potentially low recall) to provide positive examples even in absence of any other training data.
In this paper, we present and evaluate \textbf{\emph{KERMIT}} (\textbf{\emph{K}}nowl\textbf{\emph{E}}dge g\textbf{\emph{R}}aph \textbf{\emph{M}}atch\textbf{\emph{I}}ng with \textbf{\emph{T}}ransformers), a scalable knowledge graph matching system which is based on SBERT and a fine-tuned transformer component. KERMIT can match knowledge graphs with and without the provisioning of a reference alignment.

\section{Related Work}
\emph{Transformers} are deep learning architectures which combine stacked encoder layers with a self-attention~\cite{DBLP:conf/nips/VaswaniSPUJGKP17} mechanism. These architectures are typically applied in unsupervised pre-training scenarios with massive amounts of data. 
Applications of transformers for the pure ontology or knowledge graph matching task are less frequent compared to the entity matching domain. Neutel and de Boer~\cite{DBLP:conf/aaaiss/NeutelB21} use plain BERT similarity scores. They find that plain fasttext similarities still outperform SBERT descriptions; they further report that SBERT descriptions are the most useful given multiple transformer-based approaches. The work presented in this paper is similar in that it also uses SBERT. However, we use a more sophisticated pipeline, where SBERT embeddings are complemented with fine-tuned cross-encoders and further alignment repair techniques. This paper is also more comprehensive in its evaluation comprising of multiple large scale datasets. 
The MEDTO system~\cite{DBLP:conf/kdd/HaoLEQOS021} uses a graph neural network approach (GNN) to match data to medical ontologies. Due to the fact that each node in the graph needs a vector representation, they also use transformer based models to convert the concept names into such a representation. 
In 2021, the MELT
%\footnote{MELT stands for \emph{Matching Evaluation Toolkit}. It is a software framework for ontology and knowledge graph matcher development and evaluation. For more information, see~\url{https://dwslab.github.io/melt/}} 
framework~\cite{DBLP:conf/i-semantics/HertlingPP19,DBLP:conf/semweb/HertlingPP20,DBLP:conf/esws/PortischHP20} has been extended to also support a transformer filter~\cite{DBLP:conf/semweb/HertlingPP21}. This component is also used by the F-TOM~\cite{DBLP:conf/semweb/KnorrP21} matching system. Similarly to F-TOM, the TOM matching system~\cite{DBLP:conf/semweb/KossackBKP21} uses a transformer - however, TOM uses a zero-shot SBERT model rather than a fine-tuned cross-encoder. All three publications rely on a traditional blocking approach to reduce the computational complexity (at the expense of reduced recall). Moreover, TOM and the MELT component both require a sample from the reference alignment in order to fine-tune the cross-encoder. In our paper, we overcome those limitations by using a multi-stage matching pipeline.

\section{Approach}

\begin{figure}[t]
    \centering
    \includegraphics[width=0.85\textwidth]{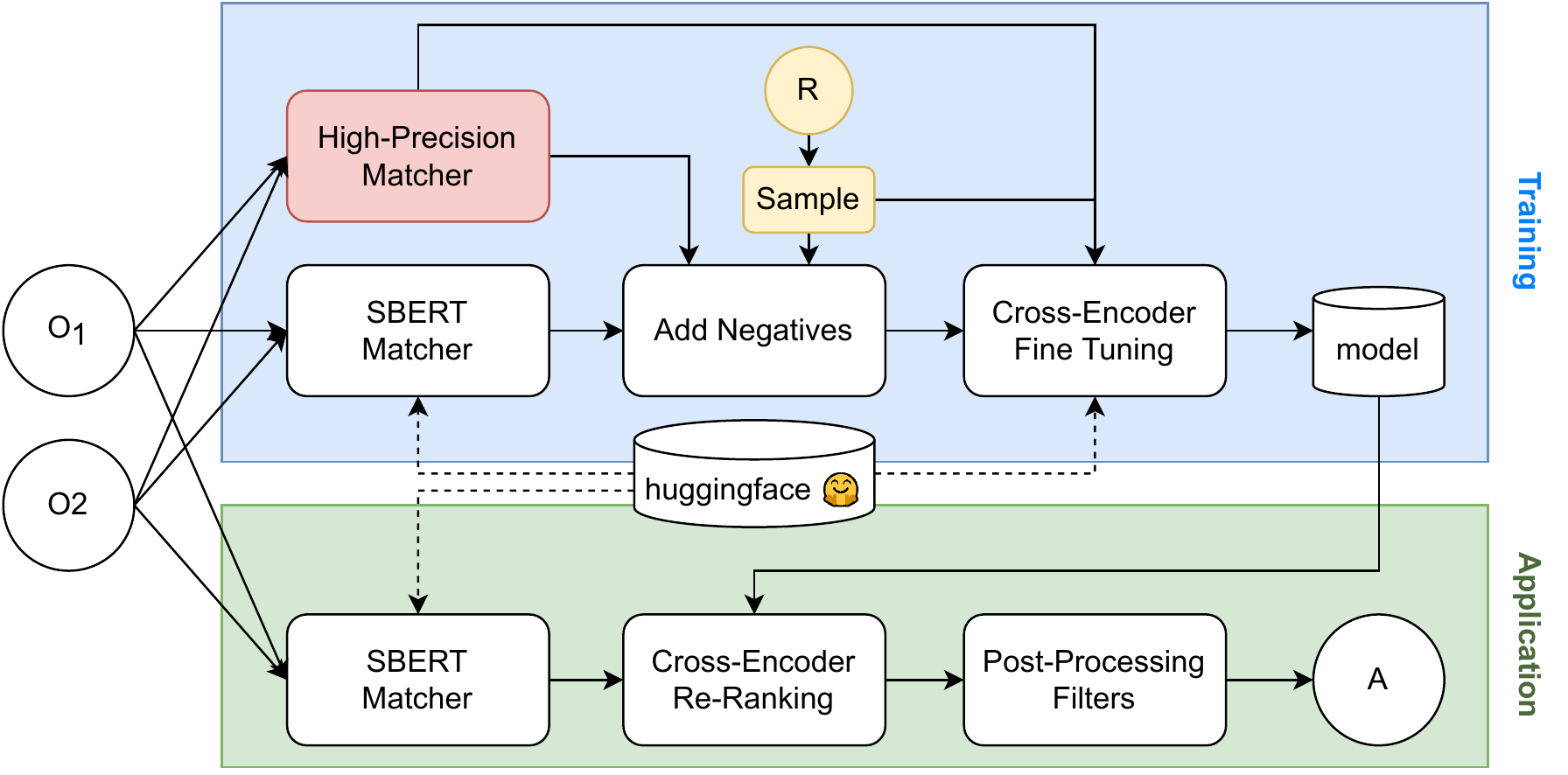}
    \caption{Overview of the approach. The red and yellow parts are alternatives.}
    \label{fig:overview}
\end{figure}

\paragraph{Matching Pipeline Overview}
We propose a multi-step pipeline. The approach is visualized in Figure~\ref{fig:overview}. In a first step, the cross encoder needs to be fine-tuned in order to recognize matches (and to discard non-matches). In Figure~\ref{fig:overview}, this step is represented in the upper blue box. Once a fine-tuned cross-encoder model is available, it can be applied in the actual matching pipeline. 

\paragraph{Training.}
In a first step, a high-recall alignment is generated (output of the SBERT matcher). For each element $e_1 \in O_1$ the top $k$ closest concepts $E_{2_k} \subset O_2$ with $|E_{2_k}| \leq k$ are added to the high-recall alignment. In this paper, $k=5$ is used. As we show in the evaluation section, if the correct SBERT model is picked, no fine-tuning is required.
In order to train the cross-encoder, positive matches and negative matches are required. For positives, KERMIT offers two options: (1) Exploration of a high-precision matcher (red in Figure~\ref{fig:overview}) and (2) Sampling from the reference (yellow in Figure~\ref{fig:overview}); both are explained in detail in Subsection~\ref{ssec:generating_candidates}.

\paragraph{Application.}
After a cross-encoder model has been fine-tuned, the actual matching step is carried out. The SBERT matcher generates an initial recall alignment. Afterwards, the cross-encoder acts as re-ranking system and reduces the set generated by the SBERT matcher by picking the best correspondence out of the $k$ correspondences generated per concept.
Thus, the resulting complexity is $O(k*(|O_1| + |O_2|))$ in addition to the cost for retrieving the top k results (compared to $O(|O_1| * |O_2|)$ for not using the bi-encoder).
KERMIT uses multiple post-processing filters which are described in detail in Subsection~\ref{ssec:post_processing}.

\subsection{Generating Candidate Correspondences with SBERT}
\label{ssec:generating_candidates}

\begin{figure}[t]
    \centering
    \includegraphics[width=0.85\textwidth]{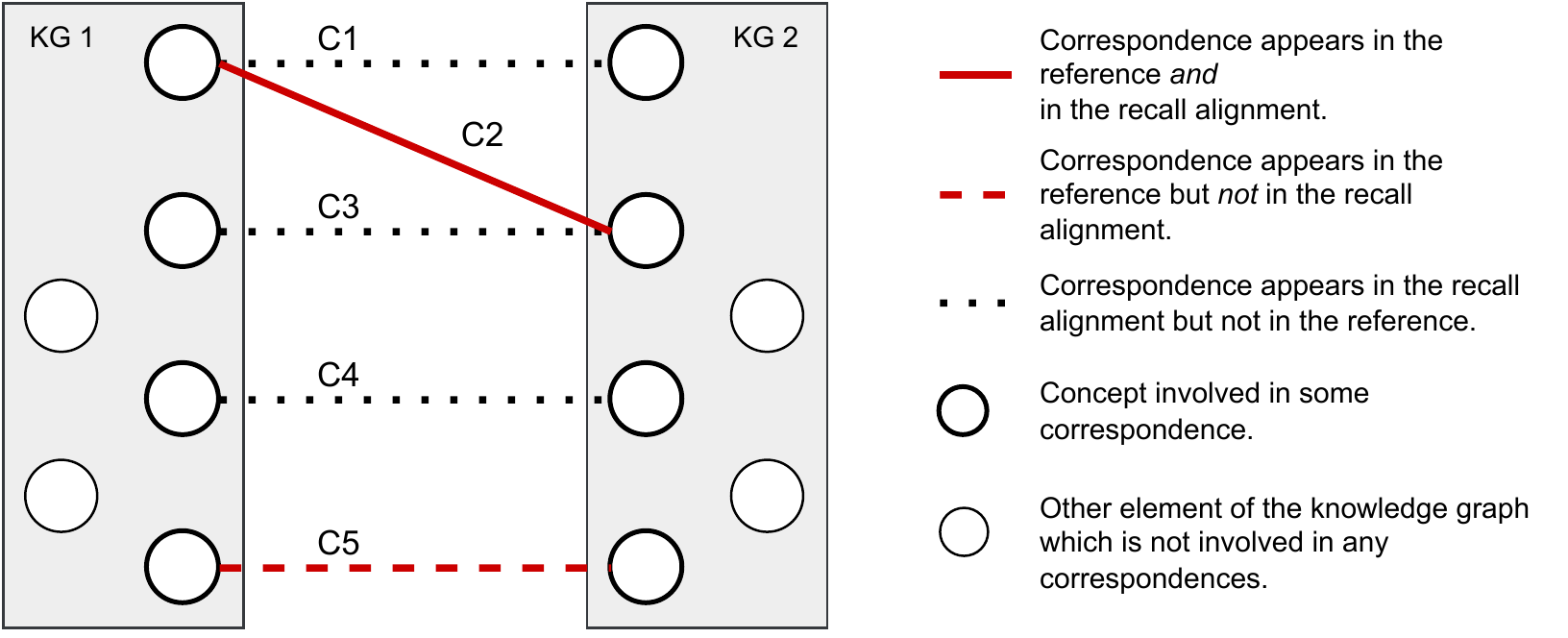}
    \caption{Generation of negatives: Given an incomplete reference and applying the one-to-one assumption, we can use C1 and C3 as negatives.}
    \label{fig:negatives_example}
\end{figure}

\paragraph{Generating Positives and Negatives.}
In order to fine-tune the cross-encoder, a set of positive and negative correspondences, more precisely text pairs, is required. 
KERMIT offers two options to obtain positives: (1) Sampling from the reference and (2) using a high-precision matcher. 
Option (1) will sample a random share $s$ from the reference alignment. In this paper, we use a constant share of $s = 20\%$.
Option (2) is applicable in situations where a reference alignment is not accessible. Any high-precision matcher can be used whose output will be considered to be correct. In this paper, we use a string based matcher which
creates correspondences for classes, properties, and instances (each in isolation). For each resource, all labels and the URI fragment are extracted and normalized (removal of camel case and non alpha numeric characters as well as lowercasing) to find matching candidates. Only entities which are mapped to only one other entity are kept.
KERMIT assumes that the one-to-one matching assumption holds. The system generates negatives using the same SBERT matcher that is also used in the application pipeline and applies the one-to-one sampling strategy: Given the high-precision alignment and the alignment produced by the SBERT matcher, it determines the wrong correspondences as correspondences where only one element is found in the high-recall alignment (but not the complete correspondence) and adds them to the training set. Note that for this approach, the high-recall alignment does not have to be complete. An example is provided in Figure~\ref{fig:negatives_example}: We can directly treat C2 and C5 as positives since they appear in the reference/high-precision alignment. Since we know that C2 must be true and that each concept can only be involved in one correspondence, we regard C1 and C3 as wrong, i.e., add them to the set of negatives. C4 is ignored, since we cannot judge whether this correspondence is true or false.
One advantage here is that the characteristics of training and test set are very similar (such as the share of positives and negatives) which is helpful for fine-tuning and using the cross-encoder. This process is visualized in the upper blue box in Figure~\ref{fig:overview}. 

\paragraph{Obtaining Textual Descriptions.} 
\label{ssec:textual_descriptions_sbert}
Concepts in ontologies and knowledge graphs may contain more than one textual description. KERMIT extracts all literals where the URI fragment of the property is either \emph{label}, \emph{name}, \emph{comment}, \emph{description}, or \emph{abstract}. This includes \texttt{rdfs:label and \texttt{rdfs:comment}}. Furthermore, the properties \texttt{pref\-Label}, \texttt{alt\-Label}, and \texttt{hiddenLabel} from the \texttt{skos} vocabulary are included. Lastly, all properties which are defined as \texttt{owl:Annotation\-Property} are followed in a recursive manner in case the object is not a label but a resource. In such a case, all annotation properties of this resource are added. All textual descriptions are collected, normalized, and duplicates are removed. The normalization is only applied to find near duplicates but the actual unmodified text is embedded.

\paragraph{Top K Calculation.} 
Textual descriptions are obtained for each $e_1 \in O_1$ and $e_2 \in O_2$. Classes, properties, and instances are embedded and searched separately such that mixed correspondences like class-instance matches are avoided. Properties are further subdivided into \texttt{owl:Object\-Property}, \texttt{owl:Data\-type\-Property}, and any other \texttt{rdf:Property} to avoid matches which are not compliant with OWL DL.
In order to generate a high-recall alignment, all textual descriptions of $O_1$ and $O_2$ are embedded using the bi-encoder (SBERT). First, all entities of the source KG are used as query and all entities of the target KG as the corpus.
For each textual description of a concept $e_1 \in O_1$, the top k closest descriptions from $O_2$ are retrieved. They are mapped to their original concepts which serve as a set of candidates to be matched to $e_1$. Those correspondences are added to the recall alignment.
The confidence is set to the similarity in the embedding space. In case multiple textual representations of two concepts are in the top k descriptions, the closest one is used. The process is repeated in the opposite direction such that each $e_2 \in O_2$ is used as query and all entities from $O_1$ as corpus. This is necessary because the operation is not symmetric and target entities may have different top k matches when they are used as a query.  

\subsection{Fine-Tuning of Cross-Encoder}

\begin{figure}[t]
    \centering
    \includegraphics[width=\textwidth]{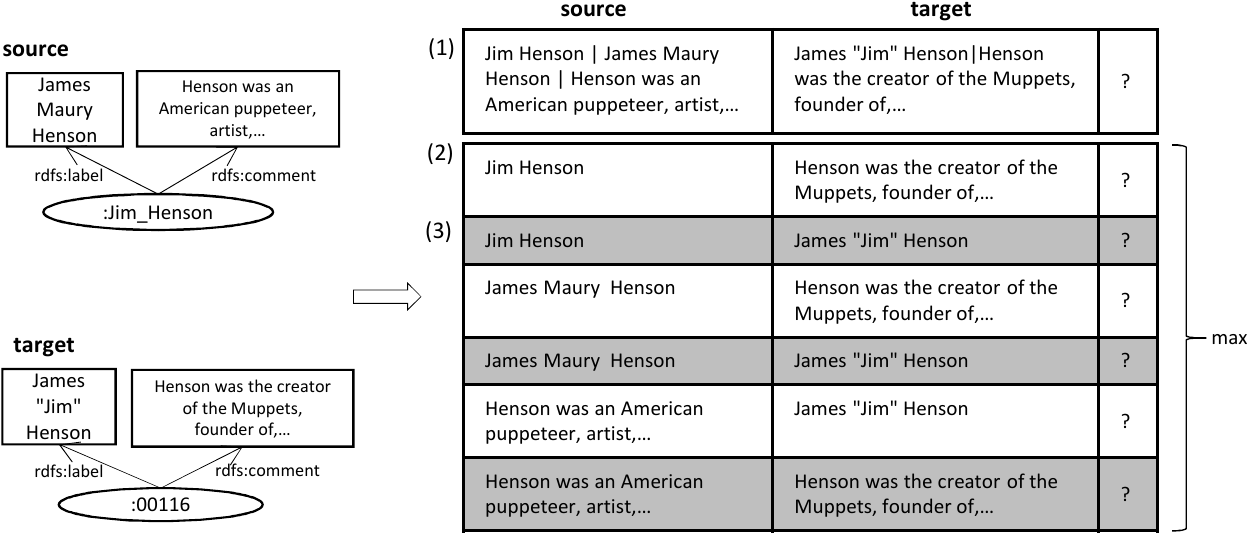}
    \caption{Obtaining textual descriptions for the fine-tuning step.}
    \label{fig:textExtraction}
\end{figure}

\paragraph{Obtaining Textual Descriptions.}
In comparison to the bi-encoder, the textual representations of a concept for the cross encoder need to be reduced as they are computed in a cross product. Thus, a slightly different approach to extract text from resources is applied. Figure~\ref{fig:textExtraction} presents three options:

\begin{enumerate}
    \item All extracted texts are concatenated together to form one input.
    \item All texts are used as own representations and each text is compared with all other extracted texts from the corresponding resource of the target KG. In the example, this corresponds to the whole lower table.
    \item The extracted texts are grouped for one resource and they will be compared only in a cross product with the same group of the other concept. Thus, we still compare individual texts 
    but reduce the overall amount of examples. In Figure~\ref{fig:textExtraction}, this corresponds to the rows with a gray background. There are only two groups: (a) short texts like URI fragments, literals of properties where the fragment is \texttt{label} or \texttt{name}, and literals which are connected with annotation properties. (b) long texts like literals of properties with URI fragment equal to \texttt{comment}, \texttt{description}, or \texttt{abstract} and the longest literal (based on the lexical representation) connected to the resource.
\end{enumerate}

\noindent Reducing the amount of examples is important because the first component in the matching pipeline (SBERT) will generate lots of correspondences (depending on the $k$ in the top $k$ retrieval step) which need to be analyzed. Furthermore, with the third approach it can be ensured that short texts are compared only with other short texts. The same logic is also applied to long texts. This reduction of examples further helps the cross-encoder to learn meaningful representations.

\paragraph{Tuning Process.}
The set of positive and negative text pairs is used to fine-tune the corresponding cross-encoder on a test case basis. The trainer class of the huggingface transformers library~\cite{DBLP:conf/emnlp/WolfDSCDMCRLFDS20} is used with default settings. If texts are too large to fit into the model, we truncate the longer of the two until both textual representations are short enough. 
The resulting dataset is highly unbalanced (due to the top 5 retrieval in the first step) 
and has a lot more negatives than positives (only one correspondence out five can be correct). During training, examples are randomly assigned to batches which results in batches without positives.
Thus, the training batch size is a crucial hyperparameter and the largest possible value is chosen. It is determined by a dataset dependent approach which sorts the input texts according to their length and runs one training step to check if everything fits on the GPU. The starting batch size is four and is iteratively multiplied by two until the memory is not sufficient anymore.

\subsection{Post-Processing Filters.}
\label{ssec:post_processing}
The cross-encoder reduces the initially obtained recall alignment. However, it still contains at least one correspondence for each concept in the ontologies. The alignment may, in addition, be incoherent since transformers are not aware of the logical constraints found in ontologies. Therefore, KERMIT uses multiple post-processing steps to obtain the final alignment. These steps are implemented as filters, i.e., they reduce the alignment. Hence, they improve the precision of the final alignment.

\noindent(1) Confidence Cut: Ideally, the cross encoder produces meaningful confidence scores $c \in [0, 1]$. These scores can be used to automatically remove low-confidence matches. As discussed earlier, the first filter removes all correspondences with a confidence $c < t$. 
For KERMIT, we use $t=0.5$, since this complements the softmax activation function used in the last transformer layer.

\noindent(2) MWB: The Max Weight Bipartite Filtering (MWB) component solves the assignment problem. It further reduces the many-to-many alignment to a one-to-one alignment. 
Due to the high amount of correspondences the Hungarian algorithm cannot be used directly; Cruz et al.~\cite{DBLP:conf/semweb/CruzAS08} provide an efficient alternative by reducing the problem to the maximum weight matching in the bipartite graph. This algorithm is re-implemented in KERMIT to output an optimized one-to-one alignment.

\noindent(3) ALCOMO: The \emph{Applying Locical Constraints on Matching Ontologies}~\cite{DBLP:phd/dnb/Meilicke11} (ALCOMO) system is an efficient alignment repair implementation which transforms the potentially incoherent alignment into a coherent one. 
For this paper, the algorithm has been re-implemented and integrated in KERMIT based on the original implementation. KERMIT uses the ALCOMO component with the PELLET reasoner and the greedy strategy to obtain a logically coherent alignment.

\paragraph{Implementation and Hardware}
KERMIT is implemented using Java and Python. The implementation is publicly available\footnote{\url{https://github.com/dwslab/melt/tree/master/examples/sentence-transformers}} as command line tool.
The evaluation has been performed using the MELT framework. It was performed on a server running Debian with 384 GB of RAM, 40 CPU cores (2.1 GHz), and 4 Nvidia Tesla V100 graphics cards.

\section{Evaluation}
KERMIT is evaluated on two different tracks by the Ontology Alignment Evaluation Initiative (OAEI): OAEI Anatomy~\cite{DBLP:conf/amia/BodenreiderHRCZ05} and OAEI LargeBio.

\subsection{Evaluation of the High Precision Matcher}
\label{ssec:eval_high_precision}
The results of the high precision matcher on the evaluation data is presented in Table~\ref{tab:sbert-zeroshot}. For the Anatomy and LargeBio track, the precision is at 99\%. Therefore, the cross-encoder needs to tolerate up to 1\% of noise in the positives of the training.

\subsection{Evaluation of SBERT Models}
In a first step, we evaluate SBERT models in terms of their ability to generate a high recall. The following models were evaluated: \texttt{all-MiniLM-L6-v2}, \texttt{paraphrase-albert-small-v2}, \texttt{paraphrase-TinyBERT-L6-v2}, \texttt{paraphrase-mp\-net-base-v2}, \texttt{paraphrase-MiniLM-L6-v2}, \texttt{paraphrase-MiniLM-L3-v2}, \texttt{all-mp\-net-base-v2}, \texttt{all-distilroberta-v5}. This set of models was generated by choosing (1) the most downloaded sentence similarity models suitable for this task, (2) the top performing models on six semantic search datasets, and (3) the best performing models on a smaller subset of the data. All of them are publicly available via the huggingface model hub. Out of these eight selected models, the first three perform best when applied to all previously discussed datasets of the OAEI. Their evaluation is presented below in more detail.

\begin{table}[t]
\footnotesize
	\begin{tabular}{|l|l|l|l|l|l|l|l|}
		\hline
		                     &                        & \multicolumn{3}{|c|}{Anatomy}  & \multicolumn{3}{|c|}{LargeBio} \\ \hline
		\textbf{k}           & \textbf{Model}         & Prec  & Rec            & $F_1$ & Prec  & Rec            & $F_1$ \\ \hline\hline
		\multirow{3}{*}{k=5} & all-MiniLM-L6-v2       & 0.066 & 0.970          & 0.124 & 0.068 & \textbf{0.954} & 0.127 \\
		                     & paraphrase-albert      & 0.065 & 0.956          & 0.122 & 0.066 & 0.943          & 0.124 \\
		                     & paraphrase-TinyBERT    & 0.066 & \textbf{0.974} & 0.124 & 0.067 & 0.948          & 0.125 \\ \hline
		\multirow{3}{*}{k=3} & all-MiniLM-L6-v2       & 0.105 & \textbf{0.963} & 0.189 & 0.104 & \textbf{0.943} & 0.188 \\
		                     & paraphrase-albert      & 0.102 & 0.948          & 0.185 & 0.102 & 0.931          & 0.185 \\
		                     & paraphrase-TinyBERT    & 0.105 & 0.962          & 0.189 & 0.104 & 0.938          & 0.187 \\ \hline
		\multirow{3}{*}{k=1} & all-MiniLM-L6-v2       & 0.307 & \textbf{0.933} & 0.462 & 0.320 & \textbf{0.894} & 0.471 \\
		                     & paraphrase-albert      & 0.298 & 0.912          & 0.449 & 0.316 & 0.886          & 0.466 \\
		                     & paraphrase-TinyBERT    & 0.302 & 0.925          & 0.455 & 0.319 & 0.891          & 0.470 \\ \hline
		-                    & Baseline Matcher       & 0.964 & 0.708          & 0.816 & 0.460 & 0.410          & 0.434 \\ \hline
		-                    & High Precision Matcher & 0.990 & 0.617          & 0.761 & 0.992 & 0.444          & 0.614 \\ \hline
	\end{tabular}
	\centering
	\caption[Performance of Zero-Shot Bi-Encoders, Baseline, and High-Precision Matcher]{Performance of zero-shot bi-encoders, baseline, and high precision matcher. The best recall per $k$ is highlighted with bold print. For the LargeBio track, macro averages are stated.}
	\label{tab:sbert-zeroshot}
\end{table}

\paragraph{Results.}
The results for $k=\{1, 3, 5\}$ are reported in Table~\ref{tab:sbert-zeroshot}. As baseline, the \texttt{SimpleStringMatcher} of the MELT framework is used.
All SBERT models achieve a remarkably high recall: If $k\geq3$, more than 90\% of the correspondences are retrieved independent of the dataset or SBERT model. Interestingly, the drop in recall when reducing $k$ is small. The overall best model independent of $k$ is \texttt{all-MiniLM-L6-v2}. The comparison with the baseline shows that each SBERT model outperforms the traditional high-recall matcher in terms of recall -- even with $k=1$. The higher precision of the baseline matcher also shows that the matcher is implemented with string-based methods.
Since the performance of the pre-trained models is sufficient for the matching task in terms of recall, we do not fine-tune those. In the following, we continue our experimentation with $k=5$. It is important to note that KERMIT would still achieve reasonable results with a lower $k$. KERMITs runtime performance is linear to $k$. Hence, $k$ can be scaled down in order to increase the runtime performance. 

\paragraph{Significance Testing.} 
Since the performance figures are still relatively close, we performed McNemar's asymptotic significance test for ontology alignments with continuity correction as described in~\cite{DBLP:journals/tkdd/MohammadiAHT18}. 
With a significance level of $\alpha = 0.05$, we find that the SBERT models do not produce statistically significantly different alignments on the Anatomy track. The only larger statistical variation in alignments occurs on the LargeBio track where roughly half of the alignments are statistically significantly different. Therefore, we conclude that combining SBERT models is not the best option to increase the recall further on most tracks; instead, $k$ should be increased.

\subsection{Evaluation of KERMIT} 
The set of experiments is continued with the best-performing SBERT model (\texttt{all-MiniLM-L6-v2}). 
The value of $k$ is set to 5. The idea is that the cross-encoder will assign an even better and more detailed confidence than the bi-encoder because it can analyze both texts (from source and target) at the same time and put the attention to the words which are important.
Similar to selecting SBERT models, the cross encoders are chosen based on the download rate of the huggingface model hub and commonly used models. The selected models are: \texttt{albert-base-v2}, \texttt{bert-base-cased}, and \texttt{roberta-base}.
The results of the complete matching pipeline are presented in Table~\ref{tab:resultCross}. The two columns \emph{High Prec Matcher} and \emph{20\% Reference} refer to the fine-tuning mode of the cross-encoder.

\begin{table}[t]
\footnotesize
	\begin{tabular}{|l|l|l|l|l|l|l|l|l|}
		\hline
		\multicolumn{3}{|l}{}                                              & \multicolumn{3}{|c|}{High Prec Matcher}                                              & \multicolumn{3}{|c|}{20\% Reference}                                                 \\
		Track                     & Test Case                    & Model   & P                          & R                          & $F_1$                      & P                          & R                          & $F_1$                      \\ \hline
		\multirow{3}{*}{Anatomy}  & \multirow{3}{*}{mouse-human} & albert  & \textbf{\underline{0.972}} & 0.699                      & 0.813                      & \textbf{0.966}             & 0.795                      & 0.872                      \\ \cline{3-9}
		                          &                              & bert    & 0.963                      & 0.718                      & 0.822                      & 0.962                      & \textbf{\underline{0.832}} & \textbf{\underline{0.892}} \\ \cline{3-9}
		                          &                              & roberta & 0.968                      & \textbf{0.725}             & \textbf{0.829}             & 0.950                      & 0.827                      & 0.884                      \\ \hline
		\multirow{9}{*}{LargeBio} & \multirow{3}{*}{fma-nci}     & albert  & \textbf{0.980}             & \textbf{\underline{0.830}} & \textbf{\underline{0.899}} & \textbf{\underline{0.986}} & 0.806                      & 0.887                      \\ \cline{3-9}
		                          &                              & bert    & 0.976                      & 0.817                      & 0.889                      & 0.957                      & 0.691                      & 0.803                      \\ \cline{3-9}
		                          &                              & roberta & 0.978                      & 0.829                      & 0.897                      & 0.984                      & \textbf{0.815} & \textbf{0.891}             \\ \cline{2-9}
		                          & \multirow{3}{*}{snomed-nci}  & albert  & \textbf{0.970}             & \textbf{0.553}             & \textbf{0.704}             & \textbf{\underline{0.971}} & 0.621                      & 0.752                      \\ \cline{3-9}
		                          &                              & bert    & 0.962                      & 0.543                      & 0.694                      & 0.953                      & 0.533                      & 0.684                      \\ \cline{3-9}
		                          &                              & roberta & 0.968                      & 0.552                      & 0.703                      & 0.967                      & \textbf{\underline{0.629}} & \textbf{\underline{0.762}} \\ \cline{2-9}
		                          & \multirow{3}{*}{fma-snomed}  & albert  & 0.939                      & 0.222                      & 0.359                      & \textbf{\underline{0.978}} & 0.666                      & 0.792                      \\ \cline{3-9}
		                          &                              & bert    & \textbf{0.940}             & 0.211                      & 0.345                      & 0.975                      & 0.632                      & 0.767                      \\ \cline{3-9}
		                          &                              & roberta & 0.927                      & \textbf{0.227}             & \textbf{0.365}             & 0.968                      & \textbf{\underline{0.716}} & \textbf{\underline{0.823}} \\ \hline
	\end{tabular}
	\centering
	\caption{Performance of KERMIT. Best precision, recall and F-measure per test case are highlighted in bold. The best performance independent of the positive example generation (high precision matcher vs. reference sampling) is additionally underlined.}
	\label{tab:resultCross}
\end{table}

On Anatomy, the results are very competitive with existing OAEI systems. It can be observed that the matching pipeline achieves a very high precision on this task. The performance increase when switching from a zero-shot approach to a reference sampling approach on this track, is between 2 and 7 percentage points -- depending on the actual cross-encoder used. The models score similar in terms of $F_1$ when using a precision matcher. Here, \texttt{roberta} achieves the overall highest score. When sampling from the reference, the variation in the performance is larger. The \texttt{bert} model achieves the highest $F_1$ of almost 90\%. Other systems scoring in this top performance range, such as AML~\cite{DBLP:conf/otm/FariaPSPCC13}, exploit domain-specific background knowledge hand-picked for this matching task. %

On LargeBio, the best scores are achieved on the fma-nci task, followed by snomed-nci, and fma-snomed. This is generally in line with typical OAEI evaluation systems. 
An interesting observation on this track is the fact that the unsupervised variants using the high precision matcher outperform the reference sampled versions on some test cases - such as fma-nci and anomed-nci. This is most likely due to the fact that the high precision matcher can generate more positives than the static 20\% sampling cut. 
In general, \texttt{albert} achieves the highest precisions with both training options. On the one hand, \texttt{roberta} is always better performing when using the reference sampling but on the other hand \texttt{albert} should be used when training examples are generated with a high precision matcher.
On fma-snomed, for instance, the reference-trained \texttt{roberta} configuration perfoms almost as good as AML, the top-notch 2021 matching system (which makes use of domain-specific background knowledge).

\section{Conclusion}
In this paper, we presented KERMIT, a knowledge graph matching system which is based on a combination of bi- and cross-encoders and reuses a logic-based alignment repair step in order to improve precision. We showed that bi-encoders are very suitable for blocking. It can be expected that they will replace traditional string-based blocking approaches in the future since (1) they can be easily configured in terms of how many candidates shall be generated, (2) they produced high-quality results, and (3) they are not biased towards pure string sequence matches.
The good results on domain-specific datasets show that the approach is particularly promising for domains where no specific knowledge sources exist and traditional matching systems fail due to missing background knowledge. 
The fine-tuned cross-encoders further helped to differentiate between true positives and false positives by re-ranking the correspondences accordingly. In comparison to other OAEI systems, KERMIT can already outperform a lot of them. Furthermore, we showed that a simple high precision matcher can also be used to generate positive correspondences -- especially in the case where the label is not the only textual information of a resource.
The bi-encoders already show good performance even when $k$ is reduced to three or one.
Thus, we plan in the future to also train a cross encoder with the same (or a suitable subset of) training data as the top performing bi-encoder. With such a pre-trained cross-encoder, there is no need for a high precision matcher or reference sampling.
As the batch size plays an important role during training of the cross-encoder, we also plan to try different sampling techniques to ensure that the batches always contain enough positives examples (also in cases of lower batch size).
Lastly, we plan to use cross-language models to tackle multilingual ontology matching.

\bibliographystyle{splncs04}
\bibliography{references}
\end{document}